\documentclass{sig-alternate}

\usepackage{subfigure} 

\usepackage{algorithm}
\usepackage{algorithmic}

\usepackage{url}
\usepackage{verbatim}

\begin{document}

\conferenceinfo{ADKDD'14, August 24,}{New York City, New York, U.S.A..}

\title{Multi-Touch Attribution Based Budget Allocation\\in Online Advertising}

\numberofauthors{3} 
\author{
\alignauthor
Sahin Cem Geyik *\\
       \affaddr{Applied Science Division}\\
       \affaddr{Turn Inc.}\\
       \affaddr{Redwood City, CA 94063}\\
       \email{sgeyik@turn.com}
\alignauthor
Abhishek Saxena *\\
       \affaddr{Applied Science Division}\\
       \affaddr{Turn Inc.}\\
       \affaddr{Redwood City, CA 94063}\\
       \email{asaxena@turn.com}
\alignauthor
Ali Dasdan\\
       \affaddr{Applied Science Division}\\
       \affaddr{Turn Inc.}\\
       \affaddr{Redwood City, CA 94063}\\
       \email{adasdan@turn.com}
}
\date{21 February 2014}

\maketitle
\begin{abstract}
\emph{Budget allocation}\let\thefootnote\relax\footnote{~~~~ \\ * The authors contributed to this work equally. \\~~~\\~~~} in online advertising deals with distributing the campaign (insertion order) level budgets to different sub-campaigns which employ different targeting criteria and may perform differently in terms of return-on-investment (ROI). In this paper, we present the efforts at Turn on how to best allocate campaign budget so that the advertiser or campaign-level ROI is maximized. To do this, it is crucial to be able to correctly determine the performance of sub-campaigns. This determination is highly related to the action-attribution problem, i.e. to be able to find out the set of ads, and hence the sub-campaigns that provided them to a user, that an action should be attributed to. For this purpose, we employ both last-touch (last ad gets all credit) and multi-touch (many ads share the credit) attribution methodologies. We present the algorithms deployed at Turn for the attribution problem, as well as their parallel implementation on the large advertiser performance datasets. We conclude the paper with our empirical comparison of last-touch and multi-touch attribution-based budget allocation in a real online advertising setting.

~~

\end{abstract}




~~~

\section{Introduction}
In online advertising, our goal is to serve the best ad for a given user in an online context. Advertisers often set constraints which affect the applicability of the ads, e.g., an advertiser might want to target only the users of a certain geographic area visiting web pages of certain types for a specific campaign. Furthermore, the objective of advertisers in general is to receive as many actions as possible utilizing different campaigns in parallel. Actions are advertiser defined and can be one of inquiring about or purchasing a product, filling out a form, visiting a certain page, etc.~\cite{klee_2012}.

An ad from an advertiser can be shown to a user on a publisher (website, mobile app etc.) only if the value for the ad \emph{impression} opportunity is high enough to win in a real-time auction~\cite{borgs_2007}. Advertisers signal their value via bids, which is calculated as the action probability given a user in a certain online context multiplied by the cost-per-action goal an advertiser wants to meet or beat. Once an advertiser, or the demand-side platform that acts on their behalf, wins the auction (i.e. submits the highest bid), it is responsible to pay the amount of the second highest bid (i.e. second-price auction). Due to this, each advertiser needs to carefully manage their \emph{budget} which dictates their capability to bid.

In this paper, we are focusing on the problem of distributing a campaign's budget to its sub-campaigns (with different targeting criteria) so that the return-on-investment (ROI, i.e. value received compared to the amount spent on advertising) is maximized, since the sub-campaigns may have different performances and spending capabilities due to their targeting. Furthermore, we will focus on the problem of \emph{action attribution} in determining a sub-campaign's performance (which helps with setting its budget), i.e. when an action is received by an advertiser, finding out the ads shown from which sub-campaign/s has/have caused that action. We examine both \emph{last-touch attribution} (LTA, i.e. a user's action is attributed to the \emph{last} ad s/he sees) and multi-touch attribution (MTA, i.e. a user's action is attributed fractionally to a subset of the ads s/he sees). The contributions of the paper can be summarized as:
\begin{itemize}
\item A budget allocation scheme that distributes money from the campaign top-level to the sub-campaigns according to their performance,
\item Examination of two action-attribution approaches to determine sub-campaign performance: last-touch and multi-touch, with an emphasis on the latter,
\item A methodology on finding multi-touch attribution of actions to sub-campaigns on large advertiser performance datasets (i.e. spending of campaigns and user data of impressions as well as the actions received), and it's efficient parallel implementation. This implementation has enabled us to process real-world online advertising datasets (tens of terabytes of user profile data, and multiple billions of virtual users) that are bigger than other published efforts dealing with multi-touch attribution so far,
\item An empirical comparison of last-touch versus multi-touch attribution based budget allocation on a real advertising sytem. To the best of our knowledge, this is the first paper to show how ROI is impacted by the choice of attribution method, and demonstrate the effect of MTA on a real-world online advertising campaign.
\end{itemize}
The rest of the paper is as follows. \S~\ref{sec:background_prev_work} will give background on both budget allocation and action-attribution in advertising domain as well as previous work in literature on these subjects. \S~\ref{sec:problem_def} will give the definition of the problem we would like to solve in this paper. We present our methodology on both budget allocation, as well as sub-campaign performance determination using both last and multi-touch action attribution schemes in \S~\ref{sec:methodology}. The implementation details of the methodology (system design as well as parallel implementation) is given in \S~\ref{sec:imp_details} which is followed by our preliminary results on different attribution methods for budget allocation given in \S~\ref{sec:results}. Finally, we conclude the paper and present some potential future work in \S~\ref{sec:conc_fut_work}. As a side note, we will be using the terms \emph{campaign} and \emph{insertion order} (IO), as well as \emph{sub-campaign} and \emph{line item} interchangeably throughout the paper. While the latter terms are more specific to online advertising domain, they are commonly used to describe a certain hierarchy within an advertiser.

\section{Background and Previous Work} \label{sec:background_prev_work}
In this section, we will give some preliminary information on the subject matter, as well as previous work in the literature.

\subsection{Budget Allocation in Online Advertising}
\begin{figure}[htb]
\centering
\includegraphics[width=2.8in]{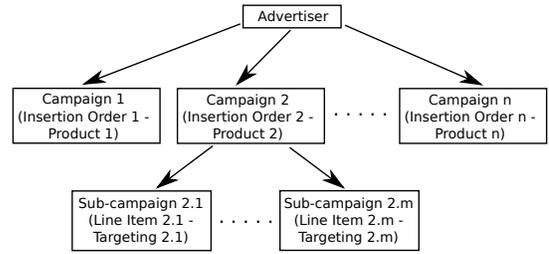}
\caption{Example of an Advertiser Hierarchy}
\label{fig:campaign_hierarchy}
\end{figure}

In online advertising, the advertisers aim to show their ad to a user on a publisher (web site, mobile app etc.), so that they get the highest number of actions for the money they spend. To be able to utilize the market more efficiently, they utilize different tactics, i.e. different campaigns with different targeting rules. For example, a sports goods company can decide to set up a campaign to show their golf equipment ads to users above a certain age or income, while their sneaker ads may be directed towards a wider audience. This inherently constructs a hierarchy for the advertisers. In our model, advertisers have different campaigns (e.g. each campaign is the advertising for a certain type of product) which we call \emph{insertion orders}, but each campaign can also have sub-campaigns (with different targeting, or different mediums (media channels), such as social, video, mobile etc.), which we call \emph{line items}. A simple example of such a hierarchy is given in Figure~\ref{fig:campaign_hierarchy}.

\begin{figure}[htb]
\centering
\includegraphics[width=2.0in]{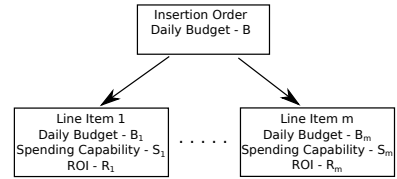}
\caption{Budget Allocation Example}
\label{fig:budget_allocation_description}
\end{figure}

\emph{Budget allocation} deals with the distribution of the daily insertion order budget to the line items under it (since we assume advertisers set up insertion order level budgets manually), and has to take into account both the spending capabilities (i.e. whether a line item's targeting allows it to reach enough users to be able to spend the money that is assigned to it), as well as performance issues (i.e. if a line item spends a certain amount of money, what is the value of actions that will be received), which is its \emph{return-on-investment} (ROI). Please see Figure~\ref{fig:budget_allocation_description} for an explanation of the budget allocation problem. In the example, the insertion order has a daily budget of $B$, and the line items are assigned daily budgets $B_i$ such that $\sum_i B_i = B$. Each line item has an ROI of $R_i$, and maximum spending capability (due to targeting, bidding etc.) of $S_i$. During budget allocation, the spending capability should be considered so that for each line item i, we have $B_i \leq S_i$ (so that no line item is assigned more money than it can spend). The overall return from the allocation given in Figure~\ref{fig:budget_allocation_description} can also be calculated as $\sum_{i} R_i ~ min(S_i,B_i)$. These calculations of course assume that we have the ROI and spending capability information, where this is not so in real settings (indeed, the main focus of this paper is learning this information). The formal problem definition (in \S~\ref{sec:problem_def}) gives further details on the budget allocation problem.

\subsection{Action-Attribution Problem in Online Advertising}
\begin{figure}[htb]
\centering
\includegraphics[width=3.2in]{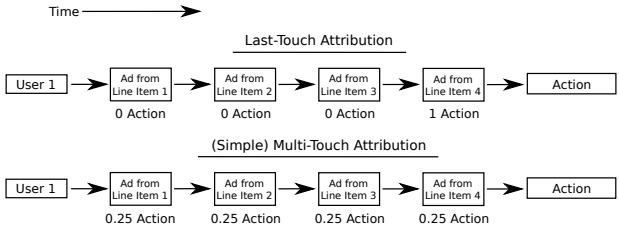}
\caption{Action Attribution Example}
\label{fig:action_att_example}
\end{figure}

As aforementioned, the aim of the advertiser is to receive as many actions as possible. Furthermore, the advertiser needs to know which sub-campaign contributed to how many actions, hence realizing the effectiveness of the different tactics utilized. The big problem for this task is the fact that the action usually happens much later than showing the ad to the user, e.g. user sees many ads online, and then purchases an item, hence it is hard to attribute actions to sub-campaigns. A very simple example for this \emph{action attribution problem} is given in Figure~\ref{fig:action_att_example}. In the example, we
present two methodologies, \emph{last-touch attribution} (the most commonly used method, attributes the action fully to the last seen ad), and \emph{multi-touch attribution} (\emph{MTA}, the action is attributed to many ads seen from the same advertiser). Please note that in the figure, we presented a very simple case of MTA, where each ad gets an equal proportion of the action, which is rarely the case in the real setting.

Naturally, action attribution and budget allocation are closely related. To be able to correctly allocate budget to sub-campaigns, we need to know how effective they are, i.e. how many actions they contributed to versus how much money was spent on them. This contribution is calculated by the action attribution methodology we employ (presented in \S~\ref{sec:methodology}).

\subsection{Previous Work}
In this section we will present some previous efforts in the literature on both budget allocation and action-attribution.

\subsubsection{Previous Efforts in Budget Allocation} \label{sec:prev_efforts_in_budget_all}
Budget allocation in the campaign level for online advertising is not a very broadly examined subject in the literature. Most of the papers so far focus on the topic of \emph{budget optimization}, i.e. given a budget constraint, how to set the bid values as well as spending profile to maximize utility (i.e. budget allocation per impression, rather than per campaign). This is significantly different than the problem we are working on, since our aim is actually to set these budget constraints. Therefore the efforts in budget optimization are complementary to our work: After we set the budgets in the campaign level, budget optimization can take place to allocate these budgets in the impression level. As an example of budget optimization, we can list \cite{archak_2010}, where the user behavior is modeled as a Markov chain. This modeling takes into account that advertising for a specific campaign type affects future behavior of the user, changing state transition probabilities. The authors model budget optimization task as a constrained optimal control problem for a Markov Decision Process (MDP).

Most of the budget allocation efforts so far aimed to maximize click revenue, since their focus stayed within the domain of search advertising. For example, the authors of \cite{ozluk_2007} propose a combined model of bid price and budget determination for keywords. They assume that click-through rate (CTR) is a function of bid price and take into account the marginal gains by increasing the bid amount, hence the budget. The solution of the optimization problem gives the optimal budget allocation. However, \cite{ozluk_2007} does not take into account the ability to deliver, which is crucial, and we focus on allocation based on actions, which is difficult due to the attribution problem. As it can be seen, due to the nature of action-based online advertising, a big portion of our discussions are to solve the attribution problem, which makes the methods based on CTR not appropriate.

In \cite{fruchter_2005}, the authors discuss the assignment of budgets to two types of search portals, generic and specialized. The authors model the allocation as an optimal control problem, and solve using dynamic programming. The biggest handicap with that approach is the assumption that the underlying parameters for earnings and clicks are known, which does not hold true and causes the methodology to be not applicable in real-world online advertising scenarios. We try to actually learn the performance of multiple sub-campaigns (which is similar to different search portals, if we take the search portal utilization as a targeting constraint) utilizing the multi-touch attribution.

The closest approach to the one proposed in this paper is given in \cite{zhang_2012}. The authors aim to do combined budget allocation and bid optimization for each campaign in an account, and employ quadratic programming method to maximize revenue. Our work differs in two ways. Firstly, \cite{zhang_2012} utilizes clicks to decide on the utility of campaigns, where we utilize actions. While clicks are straight-forward to attribute to campaigns, one of the main contributions of our work is the combined focus on attribution (which is a hard task for actions) and allocation. Again, as previously stated, any CTR-based allocation scheme is not appropriate for the domain we are focusing on. Our second difference is that we separate the budget allocation from bid optimization. The authors of \cite{zhang_2012} argue that these two should be combined since there can be well-performing keywords under overall low-performing campaigns. While such an argument is valid for search advertising, which \cite{zhang_2012} focuses on, this is not the case for online display advertising. Furthermore, due to the complicated (much more convoluted than pure keyword targeting) targeting rules involved in online display advertising campaigns, such combined optimization is often not feasible.

Finally, for a more theoretic approach, we can list \cite{alon_2012}, which focuses on the budget allocation problem to maximize the set of influenced target nodes (users). The authors model media channels (which can be taken as campaigns) and users as a bipartite graph, and the budget allocated to a media channel directly affects the number of users that are influenced by this media channel. Although this paper is not extremely relevant to ours since we aim to improve revenue (by either clicks or actions), we believe the influenced users would map nicely onto the set of buyers/clickers.

\subsubsection{Previous Efforts in Action-Attribution}
While there have been simple models utilized in the industry to perform multi-touch attribution, the first published work for data-driven allocation is given in \cite{shao_2011}. The authors provide both a bagged logistic regression model, and an intuitive probabilistic model (which uses second-order probability estimation) for attribution. 

The authors of \cite{dalessandro_2012} utilize Shapley value~\cite{shapley_1953} for attribution. It is also shown in \cite{dalessandro_2012} that the simple probabilistic scheme employed by \cite{shao_2011} is equivalent to a Shapley value formulation after rescaling, and under certain simplifying assumptions. This paper also argues that it is hard to evaluate whether one attribution of actions is better than another. Our proposed budget allocation methodology can be taken as a way to evaluate attribution methodologies, an additional contribution by our paper.

Abhishek et al.~\cite{abhishek_2013} model user behavior as a hidden Markov model (since user states are not observable, but only the outcome is, such as clicks). They later propose to utilize this behavior model to perform attribution, by attributing actions to ads that cause the user to change his/her latent state.

Finally, in \cite{wooff_2013}, the authors claim that, given no other importance information on channels, the first touch-point as well as the touch-points closer to the last one (including the last touch-point, which gets higher credit than first) get the higher credit. This attribution resembles an \emph{assymetric bathtub shape}, and the authors utilize a Beta distribution over time. Since the paper only deals with user journeys that end in action, the authors also aim at detecting the importance of initiating, intermediary, and terminating nodes for sequences within each journey, hence this way mapping channels to relevance values.

\section{Problem Definition} \label{sec:problem_def}
Let us give the formal definition of the budget allocation problem. Given the total budget $B$ for an insertion order, the set of line items $L=\{l_1,...,l_n\}$ under this IO, maximum \emph{spending capability} of each line item $S=\{S_1,...,S_n\}$, and return-on-investment (ROI) of each line item $R=\{R_1,...,R_n\}$ (the amount of dollars received by the line item, due to actions, for each dollar spent by the line item for advertising, using the specific targeting of the line item):
\[
\textrm{maximize }~ U = \sum_{i=1}^{n} R_i ~ B_i ~~ \textrm{subject to,}
\]
\[
\forall j \in [1,n] ~ B_j \leq S_j \textrm{ and } \sum_{i=1}^{n} B_i \leq B ~~.
\]
Please note that as presented in Section~\ref{sec:prev_efforts_in_budget_all}, this is significantly different than the so-called \emph{budget optimization} problem. If we have the correct values for the set $S$ and $R$, a very simple greedy approach actually optimizes the above problem:
\begin{enumerate}
\item $\textrm{B}_{\textrm{remaining}}$ = B
\item Sort line items in L according to $\textrm{R}_\textrm{i}$ (descending) into a new list $\textrm{L}_{\textrm{sorted}}$.
\item While there is budget left
\begin{itemize}
\item For each next line item $l_i$ in $\textrm{L}_{\textrm{sorted}}$
\begin{enumerate}
\item Assign $l_i$ the budget $\textrm{B}_\textrm{i}$ as min($\textrm{B}_{\textrm{remaining}},\textrm{S}_\textrm{i}$)
\item $\textrm{B}_{\textrm{remaining}} = \textrm{B}_{\textrm{remaining}} - \textrm{B}_\textrm{i}$
\item If $\textrm{B}_{\textrm{remaining}} \leq 0$, then return.
\end{enumerate}
\end{itemize}
\end{enumerate}
The problem we focus on in this paper is exactly the fact that we do not know the values $R_i$ and $S_i$ for a line item. In the next section, we show that we solve the \emph{spending capability} estimation by a simple adaptive budget assignment scheme, and \emph{return-on-investment} estimation via multi-touch attribution.

\section{Methodology} \label{sec:methodology}
As mentioned in \S~\ref{sec:problem_def}, budget allocation can be reduced to two problems: ($i$) spending capability calculation for a sub-campaign, and ($ii$) return-on-investment calculation for a sub-campaign. In this section, we will separate these two problems, and examine ways to solve them.

\subsection{Spending Capability Calculation for a Sub-Campaign}
As aforementioned, sub-campaigns (line items) apply different targeting criteria to show ads to potential buyers of a product. It is obvious that there are not the same number of users, hence the same amount of advertising budget spending capability, for all targeting criteria. We certainly do not want to assign a lot of money, no matter how high the return-on-investment may be, on a specific campaign that cannot reach enough users to be able to spend the money. It is however a hard problem to estimate exactly how much money a sub-campaign may spend, since it depends on both the reach of users, as well as the bid price (i.e. if a sub-campaign bids low, it will not be able to win ad auctions and not receive impressions, hence not be able to spend the money assigned to it). In our budget allocation approach, we apply a simple adaptive budget assignment scheme. This methodology can be summarized as follows. 
\begin{itemize}
\item If a sub-campaign is new, i.e. if we have no idea of how much it will spend, assign a learning budget that is high enough to give it a starting boost,
\item If a sub-campaign has spending data, then assign it always a bit more (e.g. increase it with a certain percentage), to explore its spending limits.
\end{itemize}
Please note that, it is possible that at any point the sum of current spending limits (calculated according to the above adaptive scheme) of sub-campaigns may be smaller than the overall campaign budget (i.e. a case of incomplete budget delivery). This usually happens if the budget assigned to a campaign is simply not possible to be spent by the sub-campaigns, hence \emph{underspend} (i.e. total spend not satisfying total budget) may occur. In the case of incomplete budget delivery, one solution that we utilize is to assign the remaining (unassigned) budget fractionally among sub-campaigns (according to their previous allocation). Although underspend may still occur, this assignment is still helpful in further calculating the spending limits of sub-campaigns, since we assign a little bit more budget to the sub-campaign than our adaptive approach suggests.

It can be seen that this simple adaptive assignment method actually tries to assign as much as possible to the sub-campaigns that perform better (high return-on-investment). This in turn tries to achieve the greedy algorithm given in \S~\ref{sec:problem_def}. Since we order the sub-campaigns/line items according to their ROI, and then assign as much as possible to the higher ranking line items, then the most important leg of the approach is calculating the ROI accurately, which is given in the next section.

\subsection{ROI Calculation for a Sub-Campaign} \label{sec:methodology:roi}
We calculate the return-on-investment for a line item as follows:
\begin{equation}
\textrm{ROI}_{l_i} = \frac{\sum_{\forall \textrm{a$_j$}} p(l_i | a_j) ~ v(a_j) }{\textrm{Money spent by } l_i} ~~~.
\label{eq:roi_formula}
\end{equation}
Above, $v(a_j)$ is the monetary value that is received by action $a_j$ (e.g. the profit that the advertiser earns by selling that specific product). In this work, we deal with CPA (cost per action) campaigns, where the advertiser provides the demand-side platform with the values of the actions that they want to receive, hence the return-on-investment is calculated as the ratio of the value of actions received to the amount of money spent for advertising. We also give the \emph{attribution} component in the above formulation by the term $p(l_i | a_j)$. This determines the percentage of the action $a_j$ that is attributed to line item $l_i$ (while for LTA, $p(l_i | a_j)$ is 0 or 1, for MTA, $p(l_i | a_j) \in [0,1]$ since we allow partial attribution of a single action to many sub-campaigns). Since the above formulation is quite straight-forward, we will focus on the attribution problem (i.e. determining $p(l_i | a_j)$) for the rest of the current section.

We have already stated that one of the most common attribution methods used is last-touch attribution, which assigns the whole action to the last ad seen by the user. In this paper, our emphasis is on multi-touch attribution, and we utilize the probabilistic model given in \cite{shao_2011}, which also originated at Turn. The methodology given in \cite{shao_2011} first calculates the empirical action probability of line items (referred to as advertising channels in the paper):
\[
p(a|l_i)=\frac{N_+(l_i)}{N_+(l_i)+N_-(l_i)}, ~ ^{*}
\]
as well as pairs of line items
\let\thefootnote\relax\footnote{\small{* As a side note, in this setting, probability of action for a sequence (regardless of the line items in it) is $p(a)=\frac{N_+}{N_+ + N_-}$, where $N_+$ is the total number of sequences (regardless of line items) that ended in action, $N_-$ is the total number of sequences that did not. This can be written in terms of action probabilities conditioned on line items as:
\[
p(a)=\sum_{S \in \{ \mathcal P \left({L}\right) - \emptyset \} } f(S) ~ p(a|S) ~ p(S)
\]
where $L$ is the set of all line items and $\mathcal P \left({L}\right)$ is the power set of (all subsets, and we further remove the empty set, $\emptyset$) $L$. $p(S)$ is the probability of a set of line items appearing together in a sequence (marginal probability of the set), which is calculated as $\frac{N_+(S)+N_-(S)}{N_+ + N_-}$, i.e. total number of sequences which have set $S$ in it, divided by the total number of sequences. $p(a|S)$ is the conditional probability of action given set $S$, and $f(S)$ is a function which gives $+1$ if set $S$ has odd number of line items in it, and $-1$ if set $S$ has even number of line items in it. This is the probability of union of conditional action events, where line items are \textbf{not} independent of each other.
}}:
\[
p(a|l_i,l_j)=\frac{N_+(l_i,l_j)}{N_+(l_i,l_j)+N_-(l_i,l_j)}.
\]
In the formulation, $N_+$ denotes the number of times that any user in the system has observed an ad sequence with an ad from line item $l_i$ (or ads from the pair of line items $l_i$ and $l_j$) that ended in action, whereas $N_-$ denotes the number of sequences that did not end in action (and had line item $l_i$, or the pair $l_i$ and $l_j$, in it). This formulation basically gives the probability that a sequence of ads shown to a user will end in conversion if it has an ad from $l_i$ (or the pair $l_i$ and $l_j$) in it. In our deployed system, we only consider actions for the last $t_{\emph{action}}$ days to be attributed to the impressions and clicks (i.e. ad sequence) that the user experienced which happened up to $t_{\textrm{association}}$ days before each action. Different values can be employed for the above two variables.

\begin{figure*}[!t]
\centering
\subfigure[First Step: Calculation of the Weights for Each Line Item]{ \includegraphics[width=5.3in]{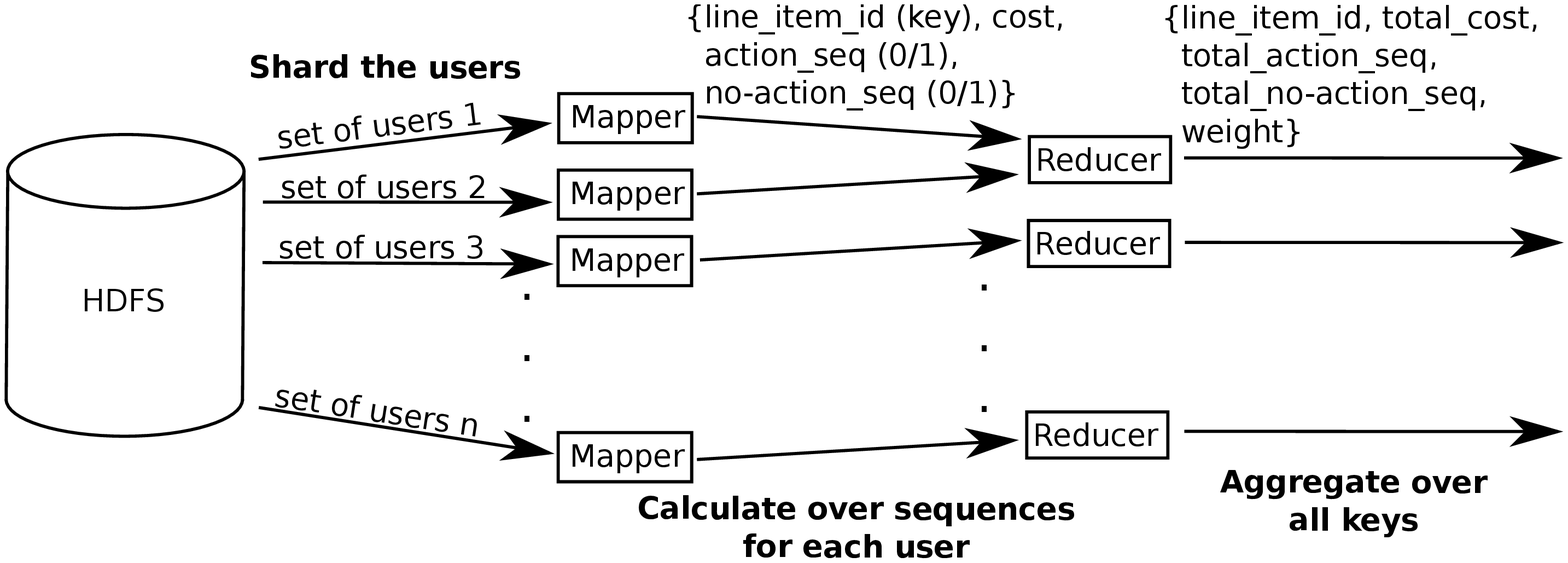} \label{subfig:mta_job_arch_1}} \\
\subfigure[Second Step: Calculation of the Attribution for Each Action, and ROI for Each Line Item]{ \includegraphics[width=5.3in]{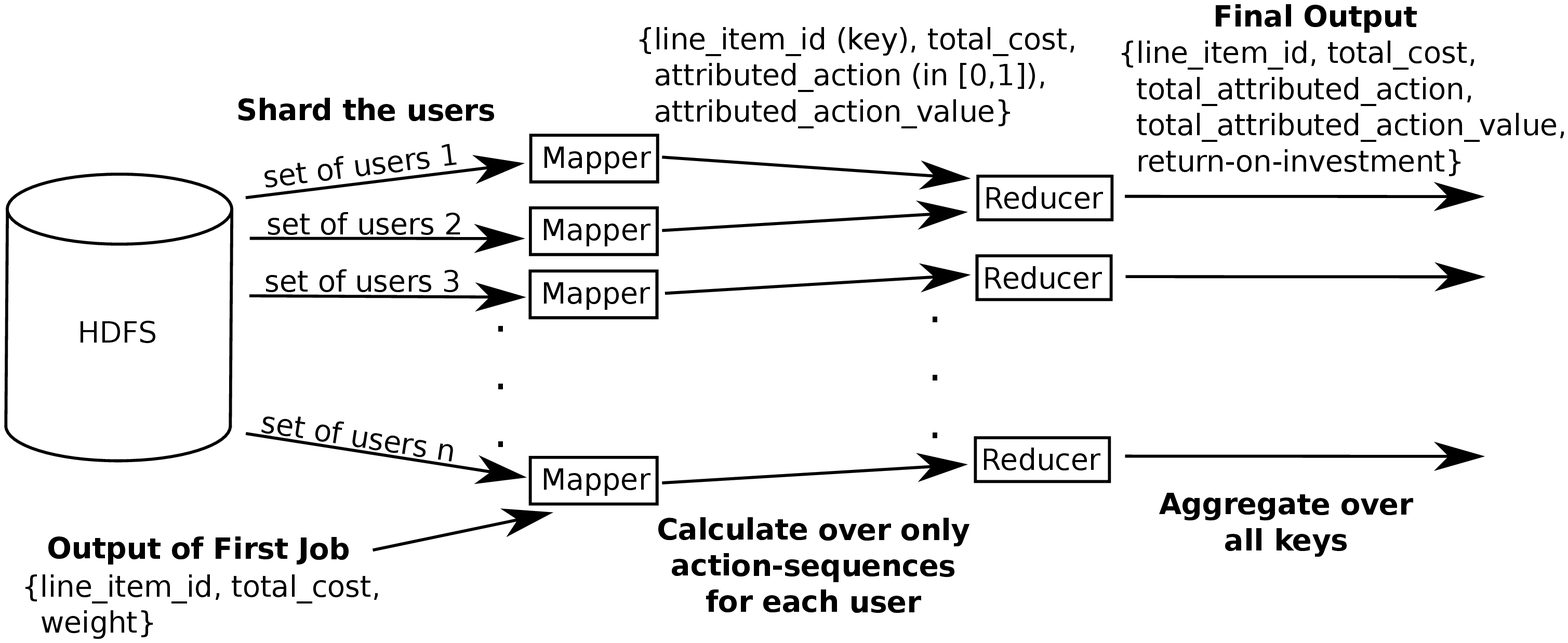} \label{subfig:mta_job_arch_2}}
\caption{Implementation Details of Employed MTA Algorithm}
\label{fig:mta_job_arch}
\end{figure*}

Once the action probabilities are calculated, the contribution weight (to be normalized to calculate actual attribution) for a line item is calculated in \cite{shao_2011} as:
\[
w(l_i)=p(a|l_i)+\frac{1}{2(N-1)}\sum_{i \neq j}\{ p(a|l_i,l_j) - p(a|l_i) - p(a|l_j) \},
\]
where N is the total number of line items under the advertiser that $l_i$ belongs to. Our experience with the current advertising system built in Turn is that the second term, i.e. the second-order calculations, does not give enough advantage in accuracy to justify the increase in processing time required to train the model (calculating the pair-wise probabilities as well as using these probabilities for the contribution weight), hence we utilize the first-order probabilities to calculate weights for the line items (although both first-order and second-order calculations are supported in our system). Therefore, the weight of each line item utilized for attribution is given as:
\begin{equation}
w(l_i)=p(a|l_i)=\frac{N_+(l_i)}{N_+(l_i)+N_-(l_i)} ~ .
\label{eq:weighting_formula}
\end{equation}

For the first step in attribution, we go through each user (i.e. web user, whose data consists of a set of impressions, clicks and actions), and only process data for a certain period (keep the actions for the last $t_{\textrm{action}}$ days, and the impressions for the last $t_{\textrm{action}} + t_{\textrm{association}}$ days, since we only attribute an action to an impression if the impression happened up to $t_{\textrm{association}}$ days before the action). Later, we extract the sequences of touch-points for the users, both those that end in an action, and those that do not. Since a sequence can have multiple touch-points from the same line item, we deduplicate those touch-points, and in the end we calculate the probability of a line item being in a sequence that ends in action as its weight (i.e. equation~\ref{eq:weighting_formula} above), which will be used for attribution  in the second step of our employed MTA algorithm. During the first step, we also calculate the amount of money spent by each line item, which is crucial to calculate ROI.

\begin{algorithm} [t!]
\caption{Second Step of Multi-Touch Attribution, Calculates the Attribution for Each Action and ROI for Each Line Item}
\label{alg:mta_second_step}
\begin{algorithmic}
\small{ 
\STATE{$t_{\emph{action}}$ = action window}
\STATE{$t_{\textrm{association}}$ = impression/click association window}
\STATE{// tp: touch-point, li: line item}

\FOR{\textbf{each} \emph{user} $u_i$}
\STATE{Keep only the imps and clicks for the time period:}
\STATE{~~~~~~~~~~~~~~~~~~~~~~ [today - $(t_{\emph{action}} + t_{\textrm{association}})$, today]}
\STATE{Keep only the actions for the time period}
\STATE{~~~~~~~~~~~~~~~~~~~~~~ [today - $t_{\emph{action}}$, today]}
\ENDFOR

\STATE{action sequence set $S_{\textrm{action}}=\emptyset$}
\STATE{// only look at action sequences}
\STATE{// since we are doing attribution}
\STATE{add each tp sequence $S_i$ that ended in action (i.e. within}
\STATE{~~~~ $t_{\textrm{association}}$ window of an action) into $S_{\textrm{action}}$}

\FOR{\textbf{each} $S_i \in S_{\textrm{action}}$}
\STATE{weightSum = $\sum_{\textrm{$l_i$ where $l_i$ has a touch-point in $S_i$}} w(l_j)$}
\FOR{\textbf{each} $l_j$ that has a touch-point in sequence $S_i$}
\STATE{actionAttributed$_{l_j}$ += $\frac{w(l_j)}{\textrm{weightSum}}$}
\STATE{totalActionAttributed$_{l_j}$ += actionAttributed$_{l_j}$}
\STATE{totalActionValue$_{l_j}$ += actionAttributed$_{l_j}$ $\times$}
\STATE{~~~~~~~~~~~~~~~~~~~~~~~~~~ valueOfActionPreceededBy$S_i$}
\ENDFOR
\ENDFOR

\FOR{\textbf{each} line item $l_j$}
\STATE{\textbf{output} totalActionAttributed$_{l_j}$ // total number of}
\STATE{~~~~~~~~~~~~~~~~~~~~~~~~~~~~~~~~~~~~~ // actions attributed to $l_j$}
\STATE{\textbf{output} totalActionValue$_{l_j}$ // total value of actions}
\STATE{~~~~~~~~~~~~~~~~~~~~~~~~~~~~~~~~~~~~ // attributed to $l_j$}
\STATE{ROI$_{l_j}$ = $\frac{\textrm{totalActionValue}_{l_j}}{\textrm{cost}_{l_j}}$}
\STATE{\textbf{output} ROI$_{l_j}$ // return-on-investment of $l_j$}
\ENDFOR
}
\end{algorithmic}
\end{algorithm}

The second step in our employed action attribution scheme is given in Algorithm \ref{alg:mta_second_step}. Since we already calculated the weights ($w(l_i)$) for the line items in the previous step, now all we have to do is to assign each action to the line items that showed at least one ad before (within a $t_{\textrm{association}}$ window) it, according to their weights (i.e. normalized weight for each line item is the fraction of the action that is attributed to it). For this purpose, we only look at the sequences that ended in action (contrary to first step, but this is needed to calculate the weights, and total cost), and in the end return the \emph{total values} of the fractional actions attributed to each line item. We also calculate ROI as given in equation \ref{eq:roi_formula} (please note that $\textrm{cost}_{l_j}$ is the total amount of money spent by line item $l_j$ for advertising, over both action and no-action sequences, and is calculated in the first step of our attribution scheme).

Please note that both of the above steps are easily parallelizable, and we present some details in the next section on how we implement our attribution and allocation system.

\section{Implementation Details} \label{sec:imp_details}
As aforementioned, the attribution scheme we employed as given in \S~\ref{sec:methodology:roi} is easily parallelizable and we have implemented the two-step algorithm on Hadoop~\cite{hadoop_2012}. This parallel implementation is necessary due to the large (multiple billions of virtual users, where each user is a set of cookies) number of users, and since we have to process the action and no-action sequences for each of them. Indeed, the amount of data we process (tens of terabytes of user profile data) is bigger than other works published so far, and represents perfectly the nature of real-world online advertising systems. The two-step MTA algorithm is run every day, for each advertiser, and is scheduled by Oozie Workflow Scheduler \cite{oozie_citation}. The current implementation at Turn takes $\approx$40 seconds per mapper for each of the first and second steps. The overall job (both steps) takes around two hours to complete every day in our production system.

The overview of our MTA implementation is given in Figure~\ref{fig:mta_job_arch}, which gives the details of the two steps separately. In Figure~\ref{subfig:mta_job_arch_1}, we present the implementation of first step in our deployed attribution algorithm, which calculates the attribution weights for each line item. The parallel processing works as follows. First, we shard the whole set of users into many mappers, which extract the action and no-action sequences, and for each sequence throws out \emph{line\_item\_id} as the key, and the following values: ($i$) \emph{cost} for the impressions (touch-points) of the line item inside the sequence, ($ii$) whether this sequence is an action sequence (0/1 value), and ($iii$) whether this sequence is a no-action sequence (0/1 value). These <key, value tuple> pairs are sent to the reducers, and the pairs with the same key end up in the same reducer which allows for aggregation. In the end, each reducer outputs the \emph{line\_item\_id} key, and the aggregated total number of action and no-action sequences which are used to calculate the weight.

The implementation of the second step of our deployed attribution scheme, where the actual action attribution as well as the line item level return-on-investment (ROI) are calculated, is presented in Figure~\ref{subfig:mta_job_arch_2}. Similar to Figure~\ref{subfig:mta_job_arch_1}, we first shard the users into mappers, and in each mapper we only go over the action sequences. Furthermore, we send the output of the first job (line item weights, as well as total costs) into the mappers, since these values are used to determine the action attribution and ROI for each line item. For each action sequence, the mappers throw out the \emph{line\_item\_id} (for each line item that had a touch-point inside this sequence that ended in an action) as key, and the following values: ($i$) total cost of line item (this is only for continuity, copied exactly from the output of first job), ($ii$) percentage of the action (that concludes this sequence) that is attributed to line item (\emph{attributed\_action} which is within the interval $[0,1]$), and ($iii$) the value of the action (that concludes this sequence) $\times$ \emph{attributed\_action} (\emph{attributed\_action\_value}), which represents the money \emph{made} by the help of advertising under this line item. Again, the same keys are collected within the same reducer, and the reducer aggregates the values to calculate the total action value (\emph{total\_attributed\_action\_value}) received by a line item, as well as the ROI for the line item (which uses both \emph{total\_attributed\_action\_value} and \emph{total\_cost} for this line item, and calculates ROI according to equation~\ref{eq:roi_formula}).

\begin{figure}[htb]
\centering
\includegraphics[width=3.2in]{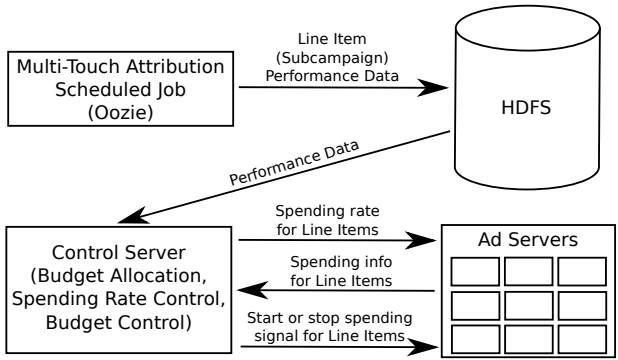}
\caption{MTA-based Budget Allocation Architecture}
\label{fig:mta_budget_alloc_architecture}
\end{figure}

The architecure we employ for MTA-based budget allocation is given in Figure~\ref{fig:mta_budget_alloc_architecture}. The budget allocation algorithm runs on the \emph{control server} which picks up the MTA-performance information from the Hadoop Distributed File System (HDFS), which is populated by the MTA Oozie job. Then, the control server calculates the daily budgets for line items, and calculates the \emph{spending rates}~\cite{klee_2013} for time periods within the day. These spending rates are sent to \emph{ad servers}, which do the spending, and send the money spent for each line item back to control server. Control server starts or stops line items from further spending (this signal is also sent to ad servers) if the line item has depleted its budget for the day.

\section{Results} \label{sec:results}
For our evaluations, we have set up two campaigns in a real online advertising environment, with the same campaign level budget, to run over 12 days within the month of November, in 2013. Both campaigns have four identical line items that run on differing targeting criteria. The only difference in the two campaigns is that the budget allocation in one is calculated utilizing the ROI values generated by MTA, and LTA in the other case. Please note that although MTA-based budget allocation is used commonly within our platform due to its advantages, we present the results of a single experiment. This is due to the fact that this kind of A/B testing requires exact set up of two campaigns to compare, hence it requires experimentation budget (i.e. money, since we assign the same amount of money to both campaigns to allocate among sub-campaigns and then spend on advertising). We are providing results in terms of \emph{return-on-investment} (ROI), \emph{effective cost per action} (eCPA) and \emph{effective cost per click} (eCPC) metrics, which are calculated in the campaign level. Our aim is to show that by allocating budgets differently to sub-campaigns according to different attribution methodologies, we improve the performance of the overall campaign. While we have explained the ROI metric throughout the paper, the latter two metrics can be described as follows:
\begin{itemize}
\item \textbf{Effective Cost per Action (eCPA):} What is the average amount of money that is spent by an advertiser (on advertising) to receive one action (i.e. purchase etc.)? This metric can be calculated as $\frac{\textrm{Advertising Cost}}{\# \textrm{ of Actions}}$.
\item \textbf{Effective Cost per Click (eCPC):} What is the average amount of money that is spent by an advertiser (on advertising) to receive one click (on its ad)? This metric can be calculated as $\frac{\textrm{Advertising Cost}}{\# \textrm{ of Clicks}}$.
\end{itemize}

\begin{figure}[htb]
\centering
\includegraphics[width=3.2in]{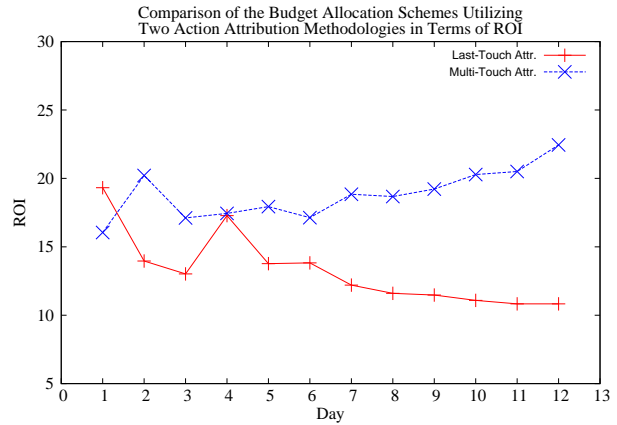}
\caption{Comparison of ROI Performance for the two budget allocation algorithms utilizing different action attribution methodologies over 12 Days. Higher ROI that has been achieved by the proposed methodology indicates better performance.}
\label{fig_ROI_comp}
\end{figure}

The results for the return-on-investment of the budget allocation applying the two attribution methodologies (LTA and MTA) is given in Figure~\ref{fig_ROI_comp}. Due to privacy issues, we
have modified the actual ROI values with a constant factor. Since we receive actions in the campaign level (i.e. when we receive an action, we know it belongs to a certain campaign, attribution to sub-campaigns comes afterwards), it is easier to calculate the overall ROI for the two identical campaigns run, to evaluate the results. It can be seen that we have much higher ROI for the MTA scheme utilized, which signifies that the ranking information (estimated ROI) is more accurate for MTA.

\begin{figure}[htb]
\centering
\includegraphics[width=3.2in]{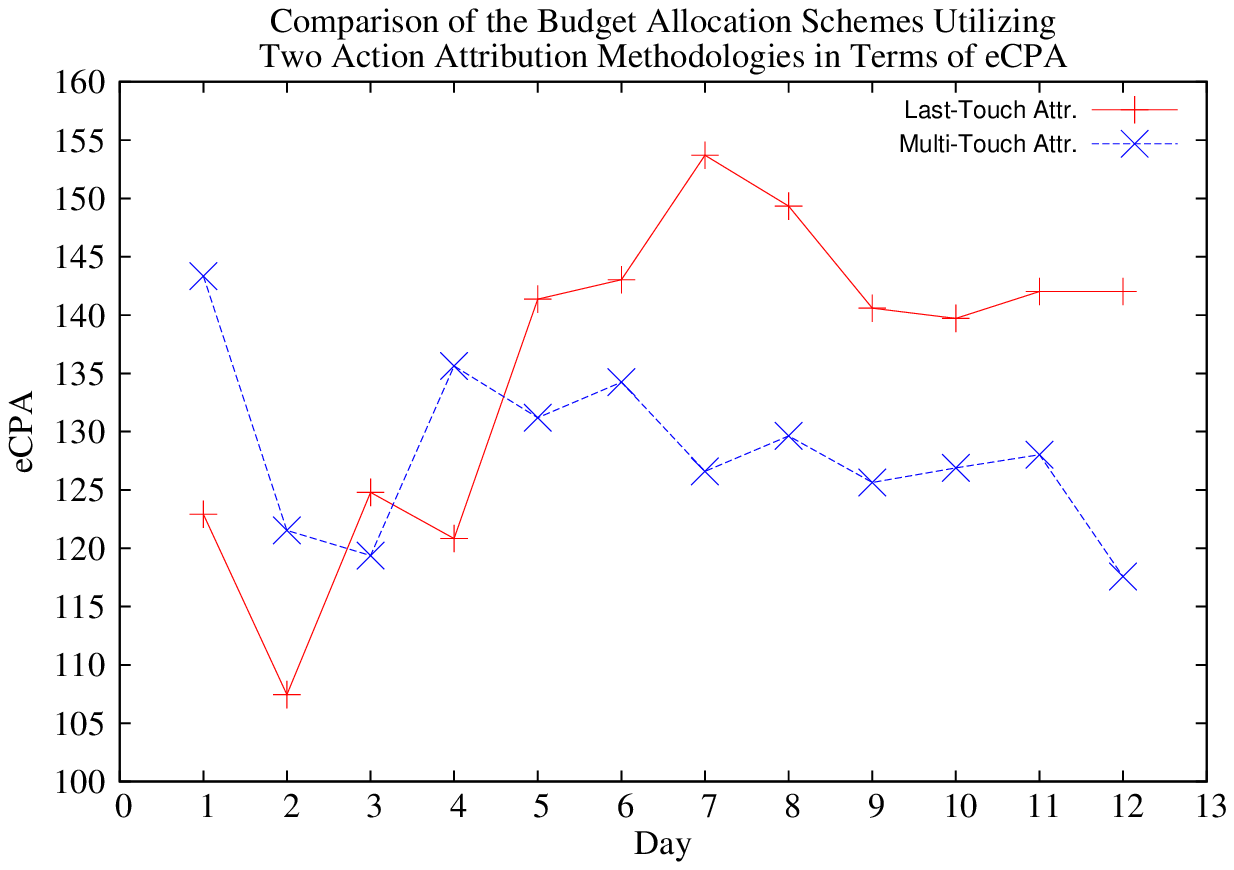}
\caption{Comparison of eCPA Performance for the two budget allocation algorithms utilizing different action attribution methodologies over 12 Days. Lower eCPA that has been achieved by the proposed methodology indicates better performance.}
\label{fig_eCPA_comp}
\end{figure}

\begin{figure}[htb]
\centering
\includegraphics[width=3.2in]{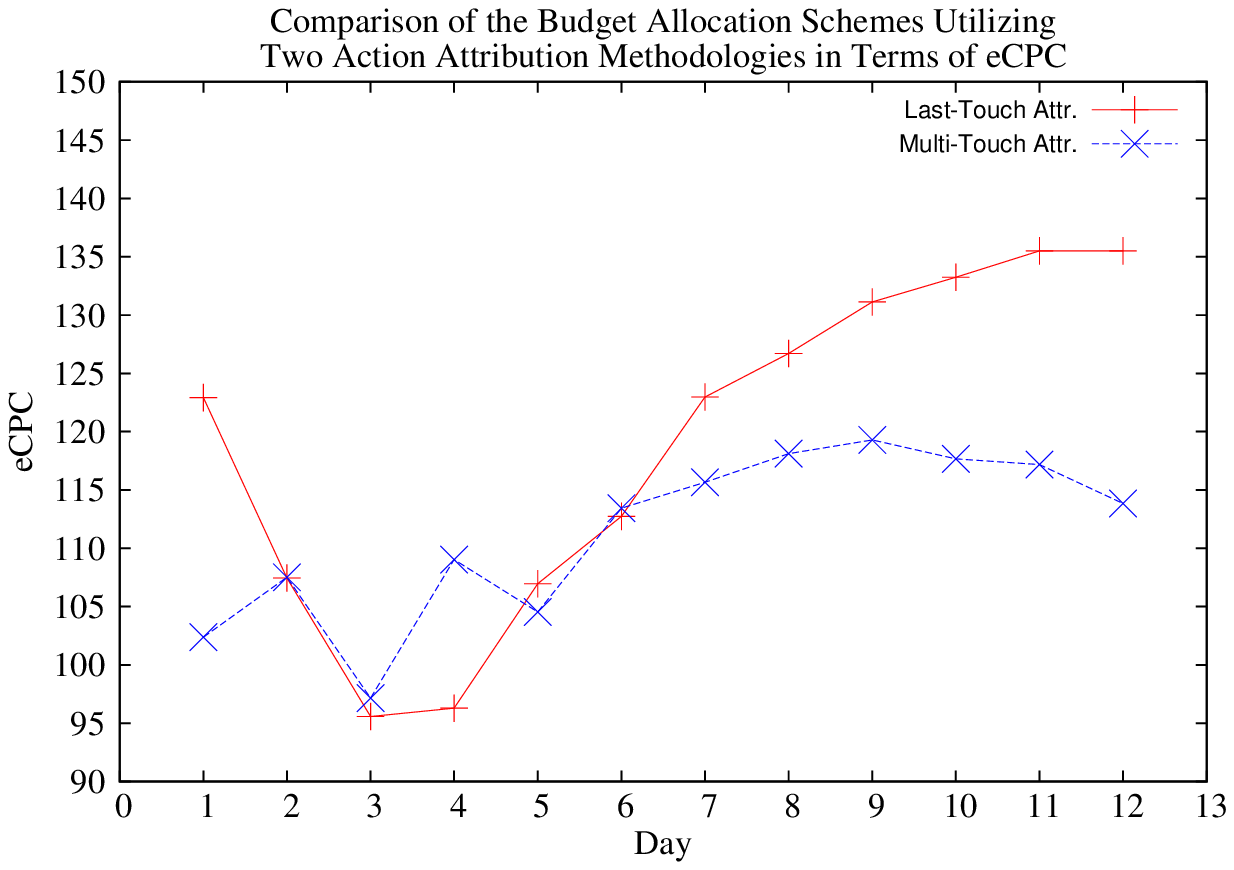}
\caption{Comparison of eCPC Performance for the two budget allocation algorithms utilizing different action attribution methodologies over 12 Days. Lower eCPC that has been achieved by the proposed methodology indicates better performance.}
\label{fig_eCPC_comp}
\end{figure}

The results in terms of eCPA and eCPC are given in Figure~\ref{fig_eCPA_comp} and Figure~\ref{fig_eCPC_comp}, respectively (again, the values are modified by a constant factor). Again, it can be seen that the budget allocation based on the MTA performs much better compared to the one that applies LTA. Please note that these eCPA and eCPC values are closely related to ROI (if the action values are the same for all actions, low eCPA means high ROI), but we see that the MTA-based allocation is much better in terms of ROI, compared to eCPA. This is due to the fact that we were able to get many more ``high quality'' (high value) actions by the MTA-based budget allocation scheme. Finally, although budget allocation was optimized towards actions via MTA, we can observe that since the MTA gives us the overall \emph{more effective} sub-campaigns, eCPC has also improved.

\begin{figure}[htb]
\centering
\includegraphics[width=1.6in]{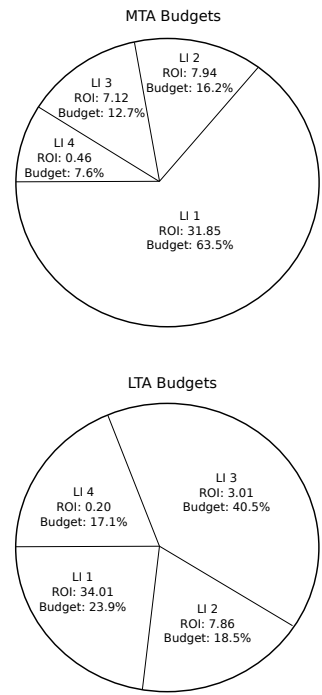}
\caption{Comparison of how the budget distributed (with the ROI received) among sub-campaigns for both budget allocation schemes. It is apparent that the MTA-based budget allocation was able to determine the ROI of campaigns with much higher accuracy and has delivered the overall budget to sub-campaigns accordingly.}
\label{fig_pie_chart}
\end{figure}

The final set of results for our experiment is given in Figure~\ref{fig_pie_chart}, which enhances our conclusion that MTA leads to better determination of sub-campaign utilities, and to improved budget allocation. In the figure, we present the percentage of the total budget allocated to each line item, alongside with the ROI received from that line item during the run of the experiment. Although we can see that the ROIs received by identical campaigns are slightly different (this difference is expected, considering different budgets are assigned), we see a remarkable correlation with the allocation achieved by the MTA-based budget allocation and the actual ROIs recorded. One more point of interest for the graph is about the highest allocated budget in the LTA case (\emph{LI 3}, i.e. line item 3). This line item is actually a retarging sub-campaign (i.e. tries to target users who have acted in some way about this product, e.g. go to the homepage, click etc.), hence it is very likely to do the last push for a user before buying a product. This of course leads to unfair assignment of actions in LTA case, unlike MTA.

\section{Conclusions and Future Work} \label{sec:conc_fut_work}
In this paper, we have focused on the problem of budget allocation in online advertising domain. We have shown that sub-campaign performance values, calculated via the multi-touch attribution, leads to better allocation of budgets. This has been demonstrated empirically in our real-world online advertising platform. We also gave a detailed explanation on the algorithms utilized for both budget allocation and multi-touch attribution, as well as their implementation.

Our future work mainly focuses on employing improved multi-touch attribution algorithms. Furthermore, we plan on the application of MTA for bidding as well, i.e. the bid is calculated utilizing the past performance values generated by the MTA algorithm.

\bibliographystyle{abbrv}
\bibliography{mta_budget}

\end{document}